\documentclass[runningheads]{llncs}
\usepackage{graphicx}
\begin{document}
\title{Building Multilingual Corpora for a Complex Named Entity Recognition and Classification Hierarchy using Wikipedia and DBpedia}
%
\titlerunning{Building Multilingual Corpora for a Complex NERC Hierarchy}
%
\author{Diego Alves\inst{1}\orcidID{0000-0001-8311-2240}\and
Gaurish Thakkar\inst{1}\orcidID{0000-0002-8119-5078} \and
Gabriel Amaral\inst{2}\orcidID{0000-0002-4482-5376}\and
Tin Kuculo\inst{3}\orcidID{0000-0001-6874-8881)}\and
Marko Tadić\inst{1}\orcidID{0000-0001-6325-820X}}
\authorrunning{Alves et al.}
%
\institute{Faculty of Humanities and Social Sciences, University of Zagreb, Zagreb 10000, Croatia \\ \email{\{dfvalio,marko.tadic\}@ffzg.hr, gthakkar@m.ffzg.hr} \and
King's College London, London, United Kingdom
\email{gabriel.amaral@kcl.ac.uk}\\
\and
L3S Research Center, Leibniz University Hannover, Hannover, Germany\\
\email{kuculo@l3s.de}}
\maketitle              
\begin{abstract}

With the ever-growing popularity of the field of NLP, the demand for datasets in low resourced-languages follows suit. Following a previously established framework, in this paper\footnote[1]{Copyright © 2021 for this paper by its authors. Use permitted under Creative Commons License Attribution 4.0 International (CC BY 4.0).}, we present the UNER dataset, a multilingual and hierarchical parallel corpus annotated for named-entities. 
We describe in detail the developed procedure necessary to create this type of dataset in any language available on Wikipedia with DBpedia information. The three-step procedure extracts entities from Wikipedia articles, links them to DBpedia, and maps the DBpedia sets of classes to the UNER labels.  This is followed by a post-processing procedure that significantly increases the number of identified entities in the final results.
The paper concludes with a statistical and qualitative analysis of the resulting dataset.


\end{abstract}

\section{Introduction}

Named entity recognition and classification (NERC) is an essential Natural Language Processing (NLP) task involved in many applications like interactive question answering, summarizing, relation extraction, and text mining. It was originally defined in the 6th Message Understanding Conference (MUC-6) as the identification of “Person”, “Location” and “Organization” types \cite{chinchor-robinson-1998-appendix}. As shown by Alves et al. \cite{alves2020uner}, NERC corpora usually respond to the specific needs of local projects and differ in terms of the complexity of the types' hierarchy and format.  

NERC systems that depend on machine learning models need to be trained with a large amount of language-specific annotated data to attain high accuracy. Nevertheless, annotating this type of data is expensive, time-consuming, and requires specific annotators, training to guarantee good inter and intra-annotator agreements \cite{lawson-etal-2010-annotating}. 


The complexity of generating NERC annotated data increases when multilingualism and a complex hierarchy of types are involved \cite{ner-sekine2004}. The Universal Named Entity Recognition (UNER) framework proposes a hierarchy composed of 3 levels\footnote[2]{https://tinyurl.com/sb3u9ve}: the first one covers eight broad categories, the second is composed of 47 named-entity types, and 15 of which are detailed in a third level with a total of 69 subtypes \cite{alves2020uner}. Therefore, our challenge was to generate automatically annotated data (silver standard) following the UNER hierarchy. 

Our goal is to parse data from Wikipedia\footnote[3]{https://www.wikipedia.org/} corpora in multiple languages, extract named entities through hyperlinks, align them with entity classes from DBpedia\footnote[4]{https://wiki.dbpedia.org/} \cite{madoc37476} and translate them into UNER types and subtypes. The idea is to propose a NERC dataset creation workflow that works for languages covered by both Wikipedia and DBpedia, including under-resourced languages. 

This paper presents the data extraction and annotation workflow and its application on English\footnote[5]{Link for English UNER Corpus v.1: https://tinyurl.com/y2taxs8b} and Croatian\footnote[6]{Link for Croatian UNER Corpus v.1: https://tinyurl.com/y4tlz4a2} languages under the licence CC BY-NC-SA 4.0. It is organized as follows: In Section 2, we present the state-of-the-art concerning NERC automatic data generation; in Section 3, we describe the details of the data extraction and annotation workflow, and in Section 4, we provide statistical analysis and qualitative evaluation of the annotation. Section 5 is dedicated to the discussion of the results, and in Section 6, we present our conclusions and possible future directions for research, including applications of the UNER corpora. 

\section{Related Work}
Due to the importance of the NERC task, the challenge of creating a quality annotated dataset has been the object of many studies. 

Yadav and Bethard \cite{vikas} and Li et al. \cite{lijing} provided complete reviews on NERC methods, however, focusing on data extraction using existing training corpora, not detailing procedures for generating new data. Nevertheless, both articles agree on the fact that a large amount of quality data is required for this task.    

To overcome the problem of cost, Lawson et al. \cite{lawson-etal-2010-annotating} proposed the usage of Amazon Mechanical Turk to annotate a specific corpus composed of e-mails. However, this method is still time-consuming and requires specific training and bonus strategies. It also relies on the availability of annotators for each specific language if the aim is to generate multilingual corpora. 

A generic method for extracting MWEs from Wikipedias was proposed in Bekavac and Tadić \cite{boke} and this covers also Multi Word Extraction named entities using local regular grammars. An automatic multilingual solution is presented by Ni et al. \cite{DBLP:journals/corr/NiDF17}. Their approach was to use annotation projection on comparable corpora or to project distributed representations of words (embeddings) from a target to a source language and, therefore, use the NER model of the source language to annotate the target one without retraining. Although with promising results, these methods require at least one quality annotated corpus (source language), which is not the case in our workflow, and that can be problematic for complex hierarchies such as UNER.  

Kim et al. \cite{kim} propose an automatic method (semi-CRF model) to label multilingual data with named entity tags using Wikipedia metadata and parallel sentences (English-foreign language) extracted from this database. This method still requires manual annotation of articles-pairs for the CRF training and the final corpus is restricted only to the encountered parallel sentences. Additionally, using Wikipedia metadata, Nothman et al. \cite{nothman} present a workflow used for the generation of silver-standard NERC corpora for English, German, Spanish, Dutch, and Russian. The idea was to transform the links between articles into new annotations by projecting the target article's classifications onto the anchor text. This method also requires previous manual annotation of a considerable number of Wikipedia articles. 

A complete automatic workflow is proposed by Weber \& Vieira \cite{vieira} for Portuguese NERC corpus generation by using Wikipedia and DBpedia information. The proposed method is the basis of our study, but we extend it to multiple languages and to a more complex NERC hierarchy. 

\section{Process for Data Extraction and Annotation}
The workflow we have developed allows the extraction of texts and metadata from Wikipedia (for any language present in this database), followed by the identification of the DBpedia classes via the hyperlinks associated with certain tokens (entities) and the translation to UNER types and sub-types (these last two steps being language independent).


Once the main process of data extraction and annotation is over, the workflow proposes post-processing steps to improve the tokenization and implement the IOB format \cite{ramshaw-marcus-1995-text}. Statistical information concerning the generated corpus is gathered, and missing UNER entities are automatically identified. 


The whole workflow is presented in detail in the project GitHub web-page\footnote[7]{https://github.com/cleopatra-itn/MIDAS} together with all scripts that have been used, and that can be applied to any other Wikipedia language.
\subsection{Components}
The following items are important components that were used for the dataset creation. These are mappers that map entities from the source to a target class or hierarchy.
\begin{enumerate}
    \item \textbf{UNER/DBpedia Mapping}: This is a mapper that connects each pertinent DBpedia class with a single UNER tag. A single extracted named entity might have more than one DBpedia class. For example, entity \textit{\textbf{2015 European Games}} have the following DBpedia classes with the respective UNER equivalences:
    \begin{itemize}
        \item \textbf{dbo:Event} -- \textit{Name-Event-Historical-Event}
        \item\textbf{dbo:SoccerTournament} -- \textit{Name-Event-Occasion-Game}
        \item \textbf{dbo:SocietalEvent} -- \textit{Name-Event-Historical-Event}
        \item \textbf{dbo:SportsEvent} --\textit{ Name-Event-Occasion-Game}
        \item \textbf{owl:Thing} -- \textit{NULL}
        
        
    \end{itemize}
    The value on the left represents a DBpedia class and its UNER equivalent is on the right side of the class. It maps all the DBpedia classes to UNER equivalent classes.
     \item \textbf{DBpedia Hierarchy}: This mapper assigns priorities to each DBpedia class. This is used to select a single DBpedia class from the collection of classes that are associated with an entity. Following are classes are their priorities.
    \begin{itemize}
                \item \textbf{dbo:Event} -- 2
        \item\textbf{dbo:SoccerTournament} -- 4
        \item \textbf{dbo:SocietalEvent} -- 2
        \item \textbf{dbo:SportsEvent} -- 4
        \item \textbf{owl:Thing} -- 1
    \end{itemize}
     For entity \textbf{\textit{2015 European Games}}, the DBpedia class  \textbf{SoccerTournament}  presides over the other classes as it has a higher priority value. If the extracted entity has two assigned classes with the same hierarchy value the first from the list is chosen as the final one. All the DBpedia classes were assigned with a hierarchy value according to DBpedia Ontology\footnote[8]{http://mappings.dbpedia.org/server/ontology/classes/}.
     
\end{enumerate}

\subsection{Main process}

The main process is schematized in the figure below and is divided into three sub-processes.

\begin{figure}
\includegraphics[width=\textwidth]{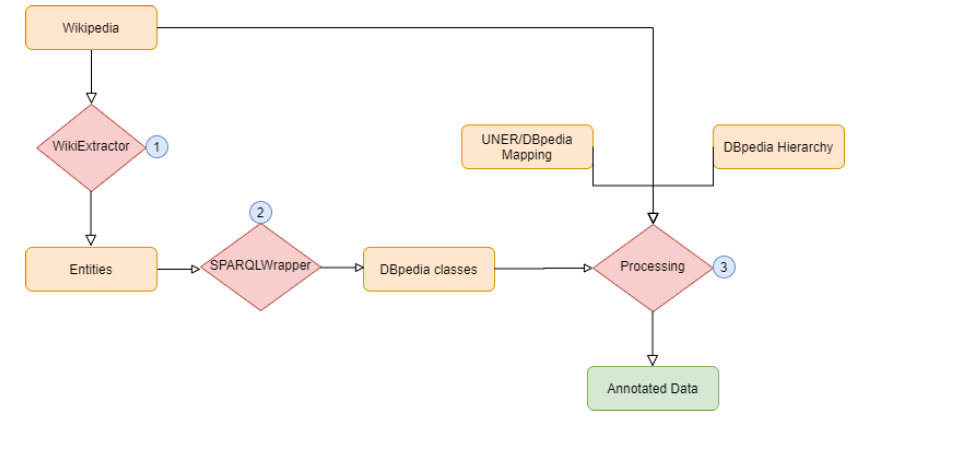}
\caption{Main process steps for Wikipedia data extraction and DBpedia/UNER annotations. Squares represent data and diamonds represent processing steps.} \label{fig1}
\end{figure}

\begin{enumerate}
    \item \textbf{Extraction from Wikipedia dumps}: For a given language, we obtain its latest dump from the Wikimedia website\footnote[9]{https://dumps.wikimedia.org/}. Next, we perform text extraction preserving the hyperlinks in the article using WikiExtractor\footnote[10]{https://github.com/attardi/wikiextractor}. These are hyperlinks to other Wikipedia pages as well as unique identifiers to those named-entities. We extract all the unique hyperlinks and sort them alphabetically. These hyperlinks will be referred to as named-entities henceforth.
    \item \textbf{Wikipedia-DBpedia entity linkin}g: For all the unique named-entities from the dumps, we query the DBpedia endpoint using a SPARQL query with SPARQLWrapper\footnote[11]{https://rdflib.dev/sparqlwrapper/} to identify the various classes associated with the entity. This step produces, for each named-entity from step 1, a set of DBpedia classes it belongs to. 
    \item \textbf{Wikipedia-DBpedia-UNER back-mapping}: For every extracted named-entity obtained in step 1, we use the set of classes produced in step 2, along with an UNER/DBpedia mapping schema, to assign UNER classes to each named-entity. For an entity, all the classes obtained from the DBpedia response are mapped to a hierarchy value, highest valued class is resolved and chosen and then it is mapped to UNER class. For constructing the final annotation dataset, we only select those sentences that have at least one single named entity. This reduces the sparsity of annotations and thus reduces the false negatives rate in our test models. This step produces an initial tagged corpus from the whole Wikipedia dump for a specific language. 
\end{enumerate}

\subsection{Post-processing steps}
The post-processing steps correspond to four different Python scripts that provide:

\begin{enumerate}
  \item The improvement of the tokenization (using regular expressions) by isolating punctuation characters that were connected with words. In addition, it applies the IOB format to the UNER annotations inside the text.
  \item The calculation of the following statistic information concerning the generated corpus: Total number of tokens, Number of Non-entity Tokens (tag “O”), Number of Entity Tokens (tags “B” or “I”) and Number of Entities (tag “B”). The script also provides a list of all UNER tags with the number of occurrences of each tag inside the corpus.
  \item Listing the entities inside the corpus (tokens and the corresponding UNER tag). Each identified entity appears once in this list, even if it has multiple occurrences in the corpus.
  \item The increment of the annotations by an automatic process using the list of entities (output of item 3 above). The script checks if, inside the corpus, entity tokens from the mentioned list were not tagged as a UNER entity. If so, the corresponding UNER tag is associated to them. Entities composed of one single character, two lower-case characters, and digits were discarded. Also, multi-tokens entities were prioritized using the longest match approach.
\end{enumerate}

Post-processing steps from 1 to 4 were applied to the UNER Croatian corpus, while for the English one, only steps 1 to 3 due to its size. Results and qualitative analysis are presented in the following section. 

\section{Data Analysis}

We have decided to test the proposed workflow with two different languages: English and Croatian. The Croatian Wikipedia \cite{wiki:Croatian_Wikipedia} is composed of 223,488 articles, while the English Wikipedia \cite{wiki:English_Wikipedia} has 6,188,204 articles (almost 28 times bigger). The aim was to check the compatibility of the process with a very well-resourced language and a low-resourced one. We have applied the main process for both languages and post-processing steps from 1-3 to the English corpus. The Croatian corpus has also been passed through the fourth step of post-processing.

\subsection{Statistical Information}

After applying the main process of the proposed workflow, we obtain, for each language, annotated text files divided into folders. The sizes of the English and Croatian UNER corpora are presented in the following table.

\begin{table}
\centering
\caption{Corpora Size.}\label{tab1}
\begin{tabular}{|l|l|l|l|}
\hline
Corpus & Total Size & Number of folders & Number of files\\
\hline
English UNER & 3.3 GB & 172 & 17,150 \\
Croatian UNER & 108 MB & 5 & 411 \\
\hline
\end{tabular}
\end{table}

After applying post-processing steps 1-3 in both Croatian and English UNER datasets, we obtain the following statistics concerning both corpora.

\begin{table}
\centering
\caption{Corpora Annotation Statistics.}\label{tab2}
\begin{tabular}{|l|l|l|}
\hline
 & English UNER Corpus & Croatian UNER Corpus\\
\hline
Total Number of Tokens & 325,395,838 & 9,388,224\\
Number of Non-Entity Tokens & 320,719,350 & 8,436,254\\
Number of Entity Tokens & 31,676,488 & 951,970\\
Number of Entities & 15,101,318 & 668,231\\
Number of Different Entities & 630,519 & 157,418\\
\hline
\end{tabular}
\end{table}

Therefore, concerning the English Corpus, 8.9\% of tokens are entities, 7.1\% in the Croatian one.

As explained previously, the UNER hierarchy is composed of categories, types, and subtypes. UNER includes the most common classes used in NERC (“Person”, “Location”, Organization”, being more detailed (subtypes):

\begin{itemize}
  \item “Person”: correspond to UNER “Name-Person-Name”
  \item “Location”: correspond to all subtypes inside the UNER types “Name-Location”.
  \item “Organization”: correspond to all subtypes inside the UNER type “Name-Organization”
\end{itemize}

Therefore, it is possible to analyse the generated corpora in terms of these more generic classes.

\begin{table}
\centering
\caption{Corpora Annotation Statistics in terms of number of occurrences of the most used NERC classes (and \% of all entities occurrences).}\label{tab2}
\begin{tabular}{|l|l|l|}
\hline
 & English UNER Corpus & Croatian UNER Corpus\\
\hline
Person & 4,200,313 (27.8\%) & 179,440 (26.8\%)\\
Location & 2,613,248 (17.3\%) & 248,801 (37.2\%)\\
Organization & 3,489,813 (23.1\%) & 74,564 (11.5\%)\\
\hline
\end{tabular}
\end{table}

These main classes correspond to 68.2\% of NEs in the English corpus and 75,5\% in the Croatian one. As explained in section 3.1, the annotation of a certain named-entity depends on the existence of hyperlinks. However, these links are not always associated with the tokens if the entity is mentioned repeatedly in the article. Therefore, to improve the corpus generated through the proposed workflow, we have established another post-processing step, which we applied to the Croatian corpus. By using the list of entities present in the corpus, we identify in the text the tokens that should have been tagged as entities, but that did not have the corresponding hyperlink.

In the following table, we present the statistics of the final post-processed Croatian corpus compared to the one obtained before this last step. 

\begin{table}
\centering
\caption{Croatian Corpus statistics after final post-processing step and delta compared to the previous corpus.}\label{tab2}
\begin{tabular}{|l|l|l|}
\hline
 & Post-processed Croatian UNER Corpus & Delta\\
\hline
Number of Non-Entity Tokens & 7,858,052 & -578,202\\
Number of Entiy Tokens & 1,530,172 & +578,202\\
Number of Entities Occurrences & 1,217,633 & +549,402 \\
\hline
\end{tabular}
\end{table}

The percentage of entity tokens is increased from 7.1\% to 16.2\%. In the next table, we focus the statistical analysis on the UNER types that can be associated with the classic NERC classes: \textit{Person}, \textit{Location} and \textit{Organisation}.

\begin{table}
\centering
\caption{Croatian UNER Corpus Statistics in terms of number of occurrences of the most used NERC classes (and \% of all entities occurrences) and delta compared to the previous corpus.}\label{tab2}
\begin{tabular}{|l|l|l|}
\hline
 & Post-Processed Croatian UNER Corpus & Delta\\
\hline
Person & 286,593 (23.5\%) & +107,153 \\
Location & 453,985 (37.3\%) & +205,184\\ 
Organization & 172,471 (14.2\%) & +97,907\\
\hline
\end{tabular}
\end{table}

\subsection{Qualitative evaluation}
The automatic annotation of the text extracted from Wikipedia requires the identification of the DBpedia classes associated with the respective tokens (via hyperlink) and the translation to UNER using (UNER/DBpedia equivalences). In order to evaluate this step, we have performed an analysis of 943 entities randomly selected concerning the English UNER Corpus. For each one, we have checked the DBpedia associated classes and the final UNER chosen tag. The next table presents the results of this evaluation.

\begin{table}
\centering
\caption{Evaluation of the annotation step: DBpedia class extraction and translation to UNER hierarchy.}\label{tab2}
\begin{tabular}{|l|l|l|}
\hline
Tag Evaluation & Number of Occurrences & Percentage\\
\hline
Correct & 797 & 85\% \\
Correct but vague & 55 & 6\%\\
Incorrect due to DBpedia & 62 & 7\%\\ 
Incorrect due to UNER association & 29 & 3\%\\
\hline
\end{tabular}
\end{table}

The analysis of this sample shows that 91\% of the entities are correctly tagged with UNER tags. However, 6\% are associated to the correct UNER type but to a generic sub-type. For example, \textbf{\textit{Bengkulu}} should be tagged as \textit{Name-Location-GPE-City} but received the tag \textit{Name-Location-GPE-GPE\_Other}.

The incorrect tags are due to errors in the DBpedia classes associated with the tokens or due to the equivalence between DBpedia and UNER rules:

\begin{itemize}
    \item \textbf{\textit{Buddhism}} is associated only to the DBpedia class \textit{EthnicGroup} and, therefore, is wrongly tagged as \textit{Name-Organization-Ethnic\_Group\_other} while it should be associated to the UNER tag \textit{Name-Product-Doctrine\_Method-Religion}.
    \item \textbf{\textit{Brit Awards}}, due to the prioritization of DBpedia class hierarchy in the choice of UNER tags, is wrongly tagged as \textit{Name-Organization-Corporation-Company} while it should receive the tag \textit{Name-Product-Award}.
\end{itemize}

It is possible to observe a considerable number of False Negative instances inside the corpora not being processed by the last post-processing step. This is due to the fact that not all entities in Wikipedia extracted articles have hyperlinks.  

A qualitative analysis of the final Croatian UNER corpus shows that the final post-processing step considerably reduces the number of False Negative instances. Nevertheless, with the actual set of rules for this final automatic annotation step, we can observe some problems concerning mostly mono-tokens entities, for example, all \textbf{\textit{jezik}} instances are associated with UNER tag \textit{Name-Product-Printing-Magazine} while in the text it may not correspond to a magazine but to the common noun \textbf{\textit{jezik}} (\textbf{\textit{language in English}}).
Also, it is important to mention that while the UNER hierarchy proposes a very complex hierarchy in terms of time and numerical expressions, this workflow covers basically UNER types and sub-types of the \textit{Name} category. 

\section{Discussion}
The implemented workflow and the generated corpora show the potential of the method presented by Weber \& Vieira \cite{vieira} comprising the extraction of texts and metadata from Wikipedia and the usage of DBpedia to annotate in terms of NERC. Our approach is broader, proposing a method that can be used for any language present in Wikipedia and using a more complex NERC hierarchy (UNER). 
Compared to other proposals of generating silver NERC standards \cite{kim}\cite{DBLP:journals/corr/NiDF17}\cite{nothman}, our workflow has the advantage of being fully automatized. However, we can identify that some improvement is needed concerning, more specifically, the last post-processing step. We have succeeded in increasing the number of annotated entities; nevertheless, the accuracy needs to be enhanced. In addition, a manual and more detailed evaluation is necessary to verify the precision and recall of the classification of the entities inside the final annotated texts. 

\section{Conclusions and Future Directions}

In this paper, we describe an automatic process for generating multilingual Named-Entity recognition corpora by using Wikipedia and DBpedia data. We also present the UNER corpus, a hierarchical named-entity corpus in Croatian and English, developed with the proposed method. We show some statistics of the corpus and detail the procedure used to create it, finishing with a qualitative evaluation. 
In our future work, we plan to extend our corpus to other under-resourced languages while evaluating our workflow's performance across the languages. Also analysing the different limitations due to the uneven content of Wikipedia across languages. Following this intrinsic evaluation of our dataset, we will train models on the obtained data to extrinsically evaluate the corpus.

\section{Acknowledgements}
The work presented in this paper has received funding from the European Union’s Horizon 2020 research and innovation program under the Marie Skłodowska-Curie grant agreement no. 812997 and under the name CLEOPATRA (Cross-lingual Event-centric Open Analytics Research Academy).

\bibliographystyle{splncs04}
\bibliography{mybibliography}

\begin{thebibliography}{10}
\providecommand{\url}[1]{\texttt{#1}}
\providecommand{\urlprefix}{URL }
\providecommand{\doi}[1]{https://doi.org/#1}

\bibitem{alves2020uner}
Alves, D., Kuculo, T., Amaral, G., Thakkar, G., Tadić, M.: Uner: Universal
  named-entity recognitionframework. In: Proceedings of the 1st International
  Workshop on Cross-lingual Event-centric Open Analytics. pp. 72--79.
  Association for Computational Linguistics (2020),
  \url{https://www.aclweb.org/anthology/W10-0712}

\bibitem{boke}
Bekavac, B., Tadić, M.: A generic method for multi word extraction from
  wikipedia. In: Proceedings of the 30th International Conference on
  Information Technology Interfaces (2008), \url{https://www.bib.irb.hr/348724}

\bibitem{chinchor-robinson-1998-appendix}
Chinchor, N., Robinson, P.: {Appendix E: MUC-7 Named Entity Task Definition
  (version 3.5)}. In: Seventh Message Understanding Conference ({MUC}-7):
  Proceedings of a Conference Held in Fairfax, Virginia, {A}pril 29 - May 1,
  1998 (1998), \url{https://www.aclweb.org/anthology/M98-1028}

\bibitem{kim}
Kim, S., Toutanova, K., Yu, H.: Multilingual named entity recognition using
  parallel data and metadata from {W}ikipedia. In: Proceedings of the 50th
  Annual Meeting of the Association for Computational Linguistics (Volume 1:
  Long Papers). pp. 694--702. Association for Computational Linguistics, Jeju
  Island, Korea (Jul 2012), \url{https://www.aclweb.org/anthology/P12-1073}

\bibitem{lawson-etal-2010-annotating}
Lawson, N., Eustice, K., Perkowitz, M., Yetisgen-Yildiz, M.: Annotating large
  email datasets for named entity recognition with {M}echanical {T}urk. In:
  Proceedings of the {NAACL} {HLT} 2010 Workshop on Creating Speech and
  Language Data with {A}mazon{'}s Mechanical Turk. pp. 71--79. Association for
  Computational Linguistics, Los Angeles (jun 2010),
  \url{https://www.aclweb.org/anthology/W10-0712}

\bibitem{madoc37476}
Lehmann, J., Isele, R., Jakob, M., Jentzsch, A., Kontokostas, D., Mendes, P.N.,
  Hellmann, S., Morsey, M., van Kleef, P., Auer, S., Bizer, C.: Dbpedia - a
  large-scale, multilingual knowledge base extracted from wikipedia. Semantic
  Web  \textbf{6}(2),  167--195 (2015). \doi{10.3233/SW-140134},
  \url{https://madoc.bib.uni-mannheim.de/37476/}

\bibitem{lijing}
li, J., Sun, A., Han, R., Li, C.: A survey on deep learning for named entity
  recognition. IEEE Transactions on Knowledge and Data Engineering
  \textbf{PP}, ~1--1 (03 2020). \doi{10.1109/TKDE.2020.2981314}

\bibitem{vieira}
Menezes, D.S., Savarese, P., Milidi{\'{u}}, R.L.: Building a massive corpus for
  named entity recognition using free open data sources  (2019),
  \url{http://arxiv.org/abs/1908.05758}

\bibitem{DBLP:journals/corr/NiDF17}
Ni, J., Dinu, G., Florian, R.: Weakly supervised cross-lingual named entity
  recognition via effective annotation and representation projection. CoRR
  \textbf{abs/1707.02483} (2017), \url{http://arxiv.org/abs/1707.02483}

\bibitem{nothman}
Nothman, J., Ringland, N., Radford, W., Murphy, T., Curran, J.R.: Learning
  multilingual named entity recognition from wikipedia. Artif. Intell.
  \textbf{194},  151--175 (2013),
  \url{http://dblp.uni-trier.de/db/journals/ai/ai194.html\#NothmanRRMC13}

\bibitem{ramshaw-marcus-1995-text}
Ramshaw, L., Marcus, M.: Text chunking using transformation-based learning. In:
  Third Workshop on Very Large Corpora (1995),
  \url{https://www.aclweb.org/anthology/W95-0107}

\bibitem{ner-sekine2004}
Sekine, S.: Named entity: History and future (2004),
  \url{http://cs.nyu.edu/~sekine/papers/NEsurvey200402.pdf}

\bibitem{wiki:Croatian_Wikipedia}
Wikipedia: {Croatian Wikipedia} --- {W}ikipedia{,} the free encyclopedia.
  \url{http://en.wikipedia.org/w/index.php?title=Croatian\%20Wikipedia\&oldid=983689658}
  (2020), [Online; accessed 10-November-2020]

\bibitem{wiki:English_Wikipedia}
Wikipedia: {English Wikipedia} --- {W}ikipedia{,} the free encyclopedia.
  \url{http://en.wikipedia.org/w/index.php?title=English\%20Wikipedia\&oldid=987449701}
  (2020), [Online; accessed 14-November-2020]

\bibitem{vikas}
Yadav, V., Bethard, S.: A survey on recent advances in named entity recognition
  from deep learning models. CoRR  \textbf{abs/1910.11470} (2019),
  \url{http://arxiv.org/abs/1910.11470}

\end{thebibliography}
%




\end{document}